\documentclass[12pt]{article}

\usepackage{epsfig}
\usepackage{amsmath,amssymb}
\usepackage{tabularx}

\newtheorem{theorem}{{\bf Theorem}}
\newtheorem{definition}[theorem]{{\bf Definition}}

\begin{document}
\setlength{\abovedisplayskip}{3pt}
\setlength{\belowdisplayskip}{3pt}

\title{Normalized Maximum Likelihood Coding for Exponential Family with Its Applications to Optimal Clustering}

\author{So Hirai
\thanks{ 
Graduate School of Information Science and Technology, 
The University of Tokyo, 
7-3-1 Hongo, Bunkyo-ku, Tokyo, JAPAN\ \  
Email: So Hirai@mist.i.u-tokyo.ac.jp\ \ 
He currently belongs to NTT DATA Corporation.}
\and Kenji Yamanishi
\thanks{
Graduate School of Information Science and Technology, 
The University of Tokyo,
7-3-1 Hongo, Bunkyo-ku, Tokyo, JAPAN \ \ 
Email: yamanishi@mist.i.u-tokyo.ac.jp}
}
\maketitle

\begin{abstract}
We are concerned with the issue of how to calculate the normalized maximum likelihood~(NML) code-length.
There is a problem that the normalization term of the NML code-length may diverge when it is continuous and unbounded and a straightforward computation of it is highly expensive when the data domain is finite .
In previous works it has been investigated how to  calculate the NML code-length for 
specific types of distributions.
We first propose a general method for computing the NML code-length for the exponential family.
Then we specifically focus on Gaussian mixture model~(GMM), and propose a new efficient method for 
computing the NML to them. We develop it by generalizing Rissanen's re-normalizing technique. 
Then we apply this method to the clustering issue, in which a clustering structure is modeled using a 
GMM, and the main task is to estimate the optimal number of clusters on the basis of the NML code-length.
We demonstrate using artificial data sets the superiority of the NML-based clustering over other criteria 
such as AIC, BIC in terms of the data size required for high accuracy rate to be achieved.
\end{abstract}


\section{Introduction}

\subsection{Motivation and Previous Works}
This paper addresses the issue of how to calculate the normalized maximum likelihood (NML) code-length for a given sequence. 
Suppose that we are given an n tuple of m-dimensional data ${\bf x}^n=({\bf x}_1,\cdots ,{\bf x}_n) \in {\cal X}^n$, 
where each ${\bf x}_i \in {\cal X} \subseteq {\mathbb R}^m$.
We define the NML distribution $f_{\rm NML}$ relative to a model class  ${\cal M}=\{f(X^n; \theta):\theta \in \Theta \} \ (n=1,2,\cdots)$ by
\begin{eqnarray}
f_{\rm NML}({\bf x}^n;{\cal M}) = \frac{f({\bf x}^n;\hat{\theta }({\bf x}^n,{\cal M}))}{{\cal C}({\cal M})},  \label{eqn:NMLdistribution} \\
{\cal C}({\cal M}) = \int f({\bf x}^n ; \hat{\theta}({\bf x}^n), {\cal M}) {\rm d}{\bf x}^n, \notag
\end{eqnarray}
where $\Theta$ is a parameter spece and $\hat{\theta}$ is a maximum likelihood estimator of $\theta$ from ${\bf x}^{n}$.
The NML code-length for ${\bf x}^n$ relative to ${\cal M}$ is calculated as follows:
\begin{eqnarray*}
-\log f_{\rm NML} ({\bf x}^n ; {\cal M})= -\log f({\bf x}^n ; \hat{\theta}({\bf x}^n, {\cal M})) + \log {\cal C}({\cal M}) ,
\end{eqnarray*}
It is known from \cite{bibli_10} that the NML code-length is optimal in the sense that it achieves the minimum of Shtarkov's minimax criterion \cite{shtarkov}.
The NML code-length is called the {\em stochastic complexity}~\cite{bibli_10} and has been 
employed as a criterion for statistical model selection on the basis of the minimum description length~(MDL) principle \cite{infcomp, Grunwald2007}.
However, there is a problem that the normalization term may diverge and 
a straightforward computation of the normalization term in the NML code-length is highly expensive.
The purpose of this paper is twofold. 
One is to propose  a method for efficient computing the NML code-length 
for the exponential family and Gaussian mixture models. The other is to demonstrate the validity of its applications to optimal clustering. 

Rissanen \cite{bibli_10} derived a formula of an asymptotic approximation of the NML code-length:
\begin{eqnarray*}
-\log p(x^{n};\hat{\theta}(x^{n}))+\frac{k}{2}\log \frac{n}{2\pi}+\log \int \sqrt{|I(\theta)|}d\theta +o(1),
\end{eqnarray*}
where $I(\theta)$ is the Fisher information matrix.
Note that this formula takes an asymptotic form. A method for exactly computing the NML code-length has been desired.
In the case where the data domain is discrete, there is a problem that the time for 
a straightforward computation of the normalization term is exponential in data size even for the 
simplest case where the class of distributions is that of mutinomial distributions.
Kontkanen and Myllym\"{a}ki proposed efficient algorithms for the NML code-length for   multinomial distributions and N\"{a}ive Bayes model \cite{petri1, petri2}. 
Meanwhile, in the case where the data domain is continuous and not bounded, 
there is a problem that the normalization term may diverge for, e.g., Gaussian distributions.
Rissanen proposed a method for circumventing this problem for
linear regression models by making an elliptic constraint for the data domain so that the normalization term does 
not diverge \cite{Denoise}.
Giurc$\breve{a}$neanu et. al. proposed another method using an rhomboid constraint \cite{Giu}.
Note that  all of these works \cite{Denoise, Giu} considered $1$-dimensional Gaussian distributions.
Hirai and Yamanishi \cite{HiraiNML} applied Rissanen's technique to the computation of the NML code-length for multi-variate Gaussian distributions.

We are specifically concerned with the applications of the NML code-length to clustering.
A mixture model may be used as a probabilistic model of clustering where each mixture component corresponds to a cluster.
The estimation of the mixture size is one of the most important issues in clustering.
Kontkanen and Myllym\"{a}ki \cite{petri2} proposed an efficient algorithm for NML-based clustering with optimal choices of mixture size for the case where the data domain was discrete.
Hirai and Yamanishi \cite{HiraiNML} proposed an algorithm for efficiently computing the NML code-length for Gaussian mixture models~(GMM) for the case where the data domain was continuous.

\subsection{Significance of This Paper}


1) {\em An extension of the computation of the NML code-length to the exponential family.}
We extend Hirai and Yamanishi's method \cite{HiraiNML} for  computing the NML code-length for Gaussian distributions and GMMs to exponential family including Gamma distributions, logistic distributions, etc.
Then we give a method for calculating the NML code-length in a general form.

2) {\em An improvement of the NML code-length for Gaussian distributions and GMMs using the renormalizing technique.}
We apply Rissanen's renormalizing technique \cite{Denoise} into Gaussian distributions and GMMs to derive new formulas for computing the NML code-lengths for them. 
Conventional formulas in \cite{HiraiNML} 
depend on the parameters by which the data domain is restricted.
The new formulas are obtained by renormalizing the likelihood with respect to the parameters, and are improved in that they are less dependent on hyper-parameters than those in \cite{HiraiNML}. 
We call the resulting code-length the renormalized maximum likelihood code-length~(RNML).
Note that the RNML are different from Rissanen's original one \cite{Denoise} in that they are derived for the case where data is multi-dimensional while Rissanen considered a specific case where it was 1-dimensional. 


3) {\em An empirical demonstration of the superiority of RNML over other criteria in the clustering scenario.}
We apply the RNML code-length to the clustering scenario in which a GMM is used as a model for clustering.
In it we employ artificial data sets to empirically demonstrate the validity of RNML in the
estimation of the number of clusters.
 We show that 
the number of clusters chosen by the RNML-based
criterion converges significantly faster to the true one than those chosen by
other criteria such as AIC, BIC, and the original NML.



\section{NML Code-Length for Exponential Family} \label{sec:NMLexpfamily}
In this section, we introduce a method of computing 
the NML code-length for the exponential family.
\vspace{-3pt}
\subsection{Exponential Family}
Below we define the  exponential family.
\begin{definition}
The probability density function belonging to the  exponential family takes the following form:
\begin{equation}
f(X ;  \theta ) = h(X)\exp \left \{ \eta(\theta)^{\rm T} T(X) - A(\eta(\theta) ) \right \}, \label{eqn:ExpFamily}
\end{equation}
where $\theta \in {\mathbb R }^D $ is a real-valued parameter vector ($D$ is the number of parameters) and $A(\eta)$ is a normalization term.
\end{definition}

The joint distribution of data ${\bf x}^n$ is given as follows:
\begin{equation*}
f({\bf x}^n ; \theta ) = \prod_{i=1}^n h({\bf x}_i)\exp \left \{ \eta(\theta)^{\rm T} T({\bf x}_i) - A(\eta(\theta) ) \right \} .
\end{equation*}
Then the maximum likelihood estimate (MLE): $\hat{\theta}({\bf x}^n)$ satisfies:

\begin{equation*}
E_{\eta(\hat{\theta}({\bf x}^n))}[T({\bf X})] = \frac{1}{n} \sum_{i=1}^n T({\bf x}_i).
\end{equation*}


\subsection{NML Code-Length for Exponential Family}


Below we consider how to calculate the normalization term: ${\cal C}({\cal M})$ as in 
(\ref{eqn:NMLdistribution}) for the exponential family.
Suppose that for any data, the MLE of $\theta $ from the data can analytically be obtained.
It is known that for the exponential family, the MLE can be calculated as a function of sufficient statistics.
Hence we may denote the MLE as follows:
\begin{eqnarray*}
\hat{\theta}({\bf x}^n) 
 &=&  \Theta \left( \frac{1}{n} \sum_{i=1}^n T({\bf x}_i) \right) ,
\end{eqnarray*}
where the $\Theta (x)$ is a certain function of $x$. 

Below we show how to calculate ${\cal C}({\cal M})$ 
by circumventing the problem that it may diverge.
The function to be integrated is expanded as follows:
\small{
\begin{eqnarray*}
f({\bf x}^n ; \theta )
&=& h({\bf x}^n | \hat{\theta}({\bf x}^n)) \times \exp \left \{ n \eta_{\theta}^{\rm T} \Theta^{-1}( \hat{\theta}({\bf x}^n) ) - n A(\eta_{\theta}) \right \} \notag \\
&=& H ({\bf x}^n | \hat{\theta}({\bf x}^n) ) \times \prod_{d=1}^D g_d (\hat{\theta}_d({\bf x}^n) | \theta ) .
\end{eqnarray*}
}\normalsize
Here we denote $\eta_{\theta} = \eta(\theta)$ and define the function $H ({\bf x}^n | \hat{\theta}({\bf x}^n) )\buildrel \rm def \over=\delta (\hat{\theta}({\bf x}^n)=\hat{\theta} )$ ($\delta(\cdot)$ is a delta function),
and the $g_d (\hat{\theta}_d({\bf x}^n) | \theta )$ is the distribution of the MLE for the $d$-th part of the parameter $\theta _{d}$. Notice here that $\theta _{d}$ is not a component of $\theta $
but rather a part of it-- a collection of components.
We assume here that parameter parts $\{\theta_{d}\}$ are independent 
with respect to $d$.
We fix $\hat{\theta}({\bf x}^n)=\hat{\theta}$ and let
\begin{equation*}
g(\hat{\theta}) \ {\buildrel \rm def \over {=}}\ \prod_{d=1}^D g_d (\hat{\theta}_d | \hat{\theta} ) .
\end{equation*}
We can calculate the normalization term ${\cal C}({\cal M})$ by integrating $g(\hat{\theta})$ with respect to $\hat{\theta}$ over the restricted domain as follows:
\begin{equation*}
{\cal C}({\cal M}) = \int_{Y(\alpha)} g(\hat{\theta}) {\rm d}\hat{\theta} ,
\end{equation*}
where we restrict the domain for the integral to be $Y(\alpha)$ where $\alpha$ is a parameter by which the integral $\hat{\theta}$ is specified.

In summary, for the exponential family,
the NML code-length can analytically be obtained 
provided that the following conditions are fulfilled:
\begin{enumerate}
\item The MLE of $\theta$ can be calculated analytically.
\item The integral of $g( \hat{\theta})$ with respect to $\hat{\theta}$ can analytically be obtained.
\end{enumerate}


\subsection{Examples}
Below we give examples of calculation of the NML code-lengths for the exponential family.
For the sake of simplicity, we focus on the normalization term ${\cal C}({\cal M})$ as in (\ref{eqn:NMLdistribution}).

\subsubsection{Gamma Distributions}
Gamma distributions belong to the exponential family.
The density function of $x^n$ for a Gamma distribution is defined as follows:
\begin{equation*}
f(x^n;k,\theta) = \prod_{i=1}^n \frac{1}{\Gamma(k)\cdot \theta^k} \cdot x_i^{k-1} \cdot \exp \left\{ -
\frac{x_i}{\theta} \right\}, 
\end{equation*}
where $k$ is a shape parameter and $\theta$ is a scale parameter.

The MLE of $\theta$ can analytically be obtained.
We consider the case where $k$ is known and fixed.
The MLE of $\theta$ is given by $\hat{\theta}(x^n) = \sum_{i=1}^n x_i / (kn)$.
Thus the joint distribution of $x^n$ is given as follows:
\begin{eqnarray*}
f(x^n;k,\theta) &=& \frac{1}{\Gamma(k)^n \cdot \theta^{k n}} \cdot \prod_{i=1}^n x_i^{k-1} \cdot \exp \left\{ -\frac{1}{\theta} \sum_{i=1}^n x_i \right\} \\
&=& H ( x^n |\ k, \hat{\theta}(x^n) ) \cdot g(\hat{\theta}(x^n);k, \theta ),
\end{eqnarray*}
where $\hat{\theta}$ is distributed according to the Gamma distribution with a shape parameter $kn$ and a scale parameter $\theta / (kn)$.
Hence $g(\hat{\theta}(x^n);k, \theta )$ is calculated as follows:
\begin{eqnarray*}
g(\hat{\theta}(x^n);k, \theta ) 
= \frac{(k n)^{k n} \cdot \hat{\theta}(x^n)^{k n-1}}{\Gamma( k n) \cdot \theta^{k n}} \cdot \exp \left\{ -\frac{k n}{\theta} \hat{\theta}(x^n) \right\} .
\end{eqnarray*}
Fix $\hat{\theta}(x^n) = \hat{\theta}$ and let $H ( x^n |\ k, \hat{\theta}(x^n) ) = \delta(\hat{\theta}(x^n) = \hat{\theta})$. 
Then we have
\begin{eqnarray*}
g(\hat{\theta};k) {\buildrel \rm def \over {=}} g(\hat{\theta};k, \hat{\theta} ) 
= \frac{(k n)^{k n}}{\Gamma(kn) \cdot {\rm e}^{k n}} \cdot \frac{1}{\hat{\theta}}.
\end{eqnarray*}
Letting hyper-parameters be $\theta_{\rm min},\ \theta_{\rm max}$ and the domain be 
\begin{eqnarray*}
Y(\theta_{\rm min}, \theta_{\rm max}) = \left\{ y^n | \theta_{\rm min} \leq \hat{\theta}(y^n) \leq \theta_{\rm max} \right\},
\end{eqnarray*}
the normalization term ${\cal C}({\cal M})$ is obtained by taking an integral of 
$g(\hat{\theta};k)$ with respect to $\hat{\theta}$ over $Y(\theta_{\rm min}, \theta_{\rm max})$as follows: 
\begin{eqnarray*}
{\cal C}({\cal M}) 
= \frac{(k n)^{k n}}{\Gamma(kn) \cdot {\rm e}^{k n}} \int_{\theta_{\rm min}}^{\theta_{\rm max}} \frac{1}{\hat{\theta}} \ {\rm d}\hat{\theta} = \frac{(k n)^{k n}}{\Gamma(kn) \cdot {\rm e}^{k n}} \log \frac{\theta_{\rm max}}{\theta_{\rm min}} .
\end{eqnarray*}
Hence, for fixed $k$, 
 we obtain a finite value of ${\cal C}({\cal M})$ for Gamma distributions.


\subsubsection{Logistic Distributions}

The logistic distributions belong to the exponential family. 
The density function of $x^n$ for a logistic distribution with a parameter $\theta $ is defined as 
\begin{equation*}
f(x^n;\theta) = \prod_{i=1}^n \frac{\theta {\rm e}^{-x_i} }{(1+{\rm e}^{-x_i})^{\theta+1}}.
\end{equation*}
The MLE of $\theta$ is analytically obtained as 
$\hat{\theta}(x^n) = n/(\sum_{i=1}^n \log (1+{\rm e}^{-x_i}))$.
Thus the joint density of $x^n$ is written as
\begin{eqnarray*}
f(x^n;\theta) &=& \theta ^n \cdot \exp \left\{ -\sum_{i=1}^n x_i - \frac{n(\theta+1)}{\hat{\theta}(x^n)} \right\} \\
&=& H ( x^n | \hat{\theta}(x^n) ) \cdot g(\hat{\theta}(x^n); \theta ),
\end{eqnarray*}
where $n/\hat{\theta}(x^n)$ is distributed according to the Gamma distribution with a shape parameter $n$ and a scale parameter $1/\theta$.
Thus $g(\hat{\theta}(x^n); \theta )$ is written as
\begin{equation*}
g(\hat{\theta}(x^n); \theta ) = \frac{\theta^n}{\Gamma(n)} \cdot \left( \frac{n}{\hat{\theta}(x^n)} \right)^{ n-1}
 \cdot \exp \left\{ -\frac{n\theta}{\hat{\theta}(x^n)} \right\} .
\end{equation*}
Fix $\hat{\theta}(x^n) = \hat{\theta}$ and let $H ( x^n | \hat{\theta}(x^n) ) = \delta(\hat{\theta}(x^n) = \hat{\theta})$. Then we have
\begin{eqnarray*}
g(\hat{\theta}) {\buildrel \rm def \over {=}} g(\hat{\theta};\hat{\theta} ) 
= \frac{n^{n-1}}{\Gamma(n) \cdot {\rm e}^{n}} \cdot \hat{\theta}.
\end{eqnarray*}
Letting $R$ be a parameter, we define the restricted domain as   
\begin{eqnarray}
Y(R) = \left\{ y^n | \hat{\theta}(y^n) \leq R \right\}. 
\end{eqnarray}
Then the normalization term ${\cal C}({\cal M})$ is obtained by taking an integral of $g(\hat{\theta})$ with respect to $\hat{\theta}$ as follows:

\begin{eqnarray*}
{\cal C}({\cal M}) 
= \frac{n^{n-1}}{\Gamma(n) \cdot {\rm e}^{n}} \int_0^{R} \hat{\theta} \ {\rm d}\hat{\theta} 
= \frac{n^{n-1}}{\Gamma(n) \cdot {\rm e}^{n}} {R}^2 .
\end{eqnarray*}
Thus we obtain the normalization term ${\cal C}({\cal M})$ that doesn't diverge.

\section{Re-normalized Maximum Likelihood} \label{sec:RNML}
We show how to compute the RNML code-length for a GMM.
Let ${\bf x}^n=({\bf x}_1,\cdots ,{\bf x}_n),\ {\bf x}_i=(x_{i1},\cdots ,x_{im})^{\rm \top} \ (i=1,\cdots ,n)$ be a given sequence where
${\bf x}_i$ is distributed according to a Gaussian distribution 
with mean ${\bf \mu}\in {\mathbb R}^m$ and variance-covariance matrix $\Sigma \in {\mathbb R}^{m\times m}$ for a some positive integer $m$ with density: 
\begin{equation*}
f({\bf x};\mu, \Sigma) = \frac{1}{(2\pi )^{\frac{m}{2}} |\Sigma |^{\frac{1}{2}}} \exp \Big \{ -\frac{1}{2}({\bf x}-\mu )^{\rm \top} \Sigma^{-1} ({\bf x}-\mu ) \Big \}.
\end{equation*}
Notice here that the normalization term in (\ref{eqn:NMLdistribution}) diverges. 
Hirai and Yamanishi \cite{HiraiNML} derived a formula of the NML distribution by restricting the range of data so that the maximum likelihood 
lies in  a bounded range specified by parameters.
It is given as follows:
\begin{equation*}
f_{\rm NML}({\bf x}^n;R,\lambda_{\rm min}) \buildrel \rm def \over = \frac{f({\bf x}^n ; \hat{\mu}({\bf x}^n),\hat{\Sigma }({\bf x}^n))}{{\cal C}(R,\lambda_{\rm min})}, 
\end{equation*}
where
\begin{eqnarray}
{\cal C}(R,\lambda_{\rm min})&=&\int_{Y(R,\lambda_{\rm min})}f({\bf y}^n;\hat{\mu}({\bf y}^n),\hat{\Sigma}({\bf y}^n)){\rm d}{\bf y}^n,\notag\\
Y(R,\lambda_{\rm min})
&{\buildrel \rm def \over {=}}& \{ {\bf y}^n| \quad || \hat{\mu}({\bf y}^n) ||^2 \leq R, \ \lambda_{\rm min}^{(j)} \leq \hat{\lambda}_j({\bf y}^n) \notag \\
&& \qquad (j=1,\cdots ,m) , \ {\bf y}^n \in {\cal X}^n  \}, \label{eqn:rangeY}
\end{eqnarray}
\vspace{-10pt}

\noindent where $R, \lambda_{\rm min}=(\lambda_{\rm min}^{(1)},\cdots ,\lambda_{\rm min}^{(m)})$ are parameters, and $\hat{\lambda}_j({\bf y}^n)$
is the $j$-th largest eigenvalue of $\hat{\Sigma}({\bf y}^n)$.
The normalization term ${\cal C}(R,\lambda_{\rm min})$ is expanded as follows \cite{HiraiNML}:
\begin{eqnarray*}
{\cal C}(R,\lambda_{\rm min}) = \frac{2^{m+1}R^{\frac{m}{2}} \prod_{j=1}^m{\lambda_{\rm min}^{(j)}}^{-\frac{m}{2}}} {m^{m+1}\Gamma (\frac{m}{2})} \times \Big ( \frac{n}{2{\rm e}} \Big )^{\frac{mn}{2}} \frac{1}{\Gamma_ m(\frac{n-1}{2})},
\end{eqnarray*}
\vspace{-5pt}

\noindent If we set the parameters: $R,\lambda_{\rm min}$ to be bounded, then the normalization term is also bounded.

Note here that the value of the normalization term depends on the choice of parameters: $R,\lambda_{\rm min}$.
Next we consider the optimization of the NML code-length with respect to 
the parameters: $R,\lambda_{\rm min}$. 
That is, we choose the optimal parameters so that they achieve the minimum of the following NML code-length:
$-\log f_{\rm NML}({\bf x}^n;R,\lambda_{\rm min})$.
The values of $R,\lambda_{\rm min}$ that make the NML code-length shortest can be considered as the maximum likelihood~(ML) estimates from ${\bf x}^n$.
The terms including $R,\lambda_{\rm min}$ in the NML code-length are given as:
\begin{equation}
\frac{m}{2} \log R - \frac{m}{2} \sum_{j=1}^m \log \lambda_{\rm min}^{(j)}. \label{eqn:NMLpara}
\end{equation}
Considering the range of parameters: (\ref{eqn:rangeY}), the ML
estimates of $R,\lambda_{\rm min}$ are given as follows:
\begin{eqnarray*}
\hat{R}({\bf y}^n) &=& || \hat{\mu}({\bf y}^n) ||^2, \\
\hat{\lambda}_{\rm min}^{(j)}({\bf y}^n) &=& \hat{\lambda}_j({\bf y}^n) \ (j=1,\cdots ,m).
\end{eqnarray*}

We then introduce hyper parameters: $\gamma =(\lambda_1,\lambda_2,R_1,R_2)$ and define the renormalized maximum likelihood (RNML) distribution by 
\begin{equation*}
f_{\rm RNML}({\bf x}^n;\gamma ) = \frac{f_{\rm NML}({\bf x}^n;\gamma, \hat{R}({\bf x}^n),\hat{\lambda}_{\rm min}({\bf x}^n))}{{\cal C}(\gamma )}, \label{eqn:RNML}
\end{equation*}
where the normalization term is expanded as follows:
\begin{eqnarray*}
 {\cal C}(\gamma ) &=& \int_{Y(\gamma)} f_{\rm NML}({\bf y}^n;\gamma, \hat{R}({\bf y}^n),\hat{\lambda}_{\rm min}({\bf y}^n)) {\rm d}{\bf y}^n, \label{eqn:Cgamma} \\
 Y(\gamma) &=& \{{\bf y}^n | \ V(\sqrt{R_1}) \leq V(\sqrt{ \hat{R}({\bf y}^n)}) \leq V(\sqrt{R_2}), \notag \\
&& \lambda_1 \leq \hat{\lambda}_{\rm min}^{(j)}({\bf y}^n) \leq \lambda_2 \ (j=1,\cdots ,m), \ {\bf y}^n \in {\cal X}^n \}, 
\end{eqnarray*}
\vspace{-10pt}

\noindent where $V(r)=2 \pi^{\frac{m}{2}}r^m /(m \Gamma (\frac{m}{2})) $, which denotes the volume of the $m$-dimensional ball with radius $r$.

The normalization term ${\cal C}(\gamma)$ is rewritten as
\begin{equation*}
{\cal C}(\gamma) = \Big( \frac{m}{2} \Big)^{m+1} \cdot \log \frac{R_2}{R_1} \cdot \Big( \log \frac{\lambda_2}{\lambda_1} \Big)^m .
\end{equation*}
%

The terms including the hyper-parameters $R_{1},R_{2},\lambda _{1},\lambda _{2}$ in the RNML code-length are given by
\begin{equation*}
\log \log \frac{R_2}{R_1} + m \log \log \frac{\lambda_2}{\lambda_1},
\end{equation*}
while those including the parameters $R,\lambda_{\rm min}^{(j)}$ in the NML code-length are given by (\ref{eqn:NMLpara}).
Comparing them each other, we see that the dependency of the RNML code-length on the hyper parameters
is lower than that of the NML code-length on the parameters by logarithmic order.

We further give a new formula of the RNML code-length relative to 
a GMM. 
\begin{theorem}
The RNML code-length of $x^{n}$ relative to a GMM is expanded as follows:
\small{
\begin{eqnarray*}
&& \ell_{\rm RNML}({\bf x}^n,z^n ; \gamma, K ) \\
&&\quad = -\log f({\bf x}^n,z^n ; K , \hat{\mu}({\bf x}^n,z^n),\hat{\Sigma }({\bf x}^n,z^n)) + \log {\cal C}_1(K,n) \\
&&\qquad \quad + \log {\cal C}_2(K,n) + \log B({\bf x}^n,z^n) + K\log I(m,\gamma ),
\end{eqnarray*}
}\normalsize
where
\small{
\begin{eqnarray}
{\cal C}_1(K,n) &=& \sum_{h_1+\cdots +h_K=n} \frac{n!}{h_1!\cdots h_K!} \prod_{k=1}^K \Big ( \frac{h_k}{n} \Big ) ^{h_k}, \label{eqn:C1} \\
{\cal C}_2(K,n) &=& \sum_{h_1+\cdots +h_K=n} \frac{n!}{h_1!\cdots h_K!} \prod_{k=1}^K \Big ( \frac{h_k}{n} \Big )^{h_k} \cdot J(h_k), \notag \\ \label{eqn:C2} \\
B({\bf x}^n,z^n) &=& \prod_{p=1}^K \frac{2^{m+1} \cdot ||\hat{\mu}_p({\bf x}^n,z^n) ||^m \cdot |\hat{\Sigma }_p({\bf x}^n,z^n)|^{-\frac{m}{2}}}{m^{m+1}\Gamma (\frac{m}{2})}, \notag \\
I(m,\gamma ) &=& {\cal C}(\gamma )=\Big( \frac{m}{2} \Big)^{m+1} \cdot \log \frac{R_2}{R_1} \cdot \Big( \log \frac{\lambda_2}{\lambda_1} \Big)^m , \notag \\
J(h_k) &=& \Big ( \frac{h_k}{2{\rm e}} \Big )^{m h_k} \cdot \frac{1}{\Gamma_m (\frac{h_k-1}{2})}. \label{eqn:J} 
\end{eqnarray}
}\normalsize
Here $h_k$ denotes the number of data belonging to the $k$-th cluster, and $\hat{\mu}_p, \hat{\Sigma}_p$ denote mean and the ML estimates of the variance-covariance matrix for the $p$-th cluster. 
\end{theorem}

Note that straightforward computation of ${\cal C}_{1}(K,n)$ 
and ${\cal C}_{2}(K,n)$ as in (\ref{eqn:C1}) and (\ref{eqn:C2}) requires $O(n^{K})$ time.
Below we give methods for efficient computation of ${\cal C}_{1}(K,n)$ 
and ${\cal C}_{2}(K,n)$.
As for the computation of ${\cal C}_{1}(K,n)$, Kontkanen and Myllym\"{a}ki 
proved the following theorem:
\begin{theorem}{\rm \cite{petri1}}
${\cal C}_{1}(K, n)$ satisfies the recursive formula:
\begin{equation}
{\cal C}_{1}(K+2, n)=C_{1}(K+1, n)+\frac{n}{K}C_{1}(K,n).
\end{equation}
Hence ${\cal C}_{1}(K,n)$ is computed in time $O(n+K)$.
\end{theorem}

As for the computation of ${\cal C}_{2}(K,n)$, we newly give the following result:
\begin{theorem}
${\cal C}_{2}(K, n)$ satisfies the following formula:
\begin{equation}
{\cal C}_{2}(K+1, n)=\sum _{r_{1}+r_{2}=n}\ _{n}C_{r_{1}}\left( \frac{r_{1}}{n}\right)^{r_{1}}\left( \frac{r_{2}}{n}\right)^{r_{2}}C_{2}(K, r_{1})J(r_{2}),
\end{equation}
where $J(r_{2})$ is as in (\ref{eqn:J}).
Hence ${\cal C}_{2}(K,n)$ is computed in time $O(n^{2}K)$.
\end{theorem}

Combining all of the theorem as above, we see that the RNML code-length of $x^{n}$
relative to a GMM is computed in time $O(n^{2}K)$.


\section{Experimental Results}
\subsection{Comparison with AIC and BIC} \label{subsec:ExAICBIC}
This section gives experimental results showing the validity of the RNML for GMMs.
We generated a number of data sequences of size $n$ according to the true GMM ${\cal M}$ of mixture size 
$K$.
Each mixture component is a Gaussian distribution with mean $\mu_k $ and variance-covariance matrix $\Sigma_k \ (k=1, \cdots ,K)$. 
For each data sequence ${\bf x}^n$ generated according to the true model ${\cal M}$, we also generated their corresponding cluster indices $z^n$ 
using the EM algorithm \cite{dlr}, where $z_{i}$ showed which cluster ${\bf x}_{i}$ came from $(i=1,\dots ,n)$. 
In our experiment, we repeated cluster generation using the EM algorithm 100 times  by changing initial values of the algorithm.
We compared  the four criteria: RNML, NML, Akaike's Information Criterion~(AIC) \cite{AIC} and Bayesian Information Criterion~(BIC) \cite{BIC} for the choice of the number of clusters. 
We calculated RNML and NML according to the method proposed in the previous sections and \cite{HiraiNML}.
We calculated AIC and BIC as follows:
\vspace{2pt}
\begin{eqnarray*}
&& AIC({\bf x}^n,z^n;K) = -2 \log f({\bf x}^n,z^n; K, \hat{\theta}({\bf x}^n,z^n)) \\
&&\qquad \qquad \qquad \qquad + m(m+3)K+K , \label{eqn:AIC_experiment} \\
&& BIC({\bf x}^n,z^n;K) = -2 \log f({\bf x}^n,z^n; K, \hat{\theta}({\bf x}^n,z^n)) \\
&& \qquad \qquad \qquad \qquad + \frac{m(m+3)K}{2}\sum _{k=1}^K\log h_k + K \log n . 
\end{eqnarray*}
\vspace{-5pt}

\noindent We measured their performance  in terms of 
the identification probability  $P(K)$ and the benefit $B(K)$ defined as follows: Letting $K$ be the true number of clusters and 
$K^*$ be the one chosen using any criterion,
\begin{eqnarray}
P(K) &=& Prob(K^*=K), \notag \\
B(K) &=& max\left\{0, 1-\frac{|K^*-K|}{T}\right\}, \label{eqn:benefit}
\end{eqnarray}
where $T$ is a given constant.
The {\em identification probability} $P(K)$ is the probability that the algorithm outputs the true number of clusters.
The {\em benefit} is a score assigned to $K$ so that if $K=K^{*}$ it takes the maximum value $1$, and it decreases linearly to zero as $|K^{*}-K|$ increases to $T$. 
The resulting benefit is calculated as the average of the benefits taken over all of random generation.
We compared RNML, AIC, and BIC in terms of how fast the identification probability and the benefit converge as sample size $n$ increases.

\begin{figure*}[!ht]
\hspace{10pt}
\begin{minipage}{.27\linewidth }
\includegraphics[width=\linewidth,clip ]{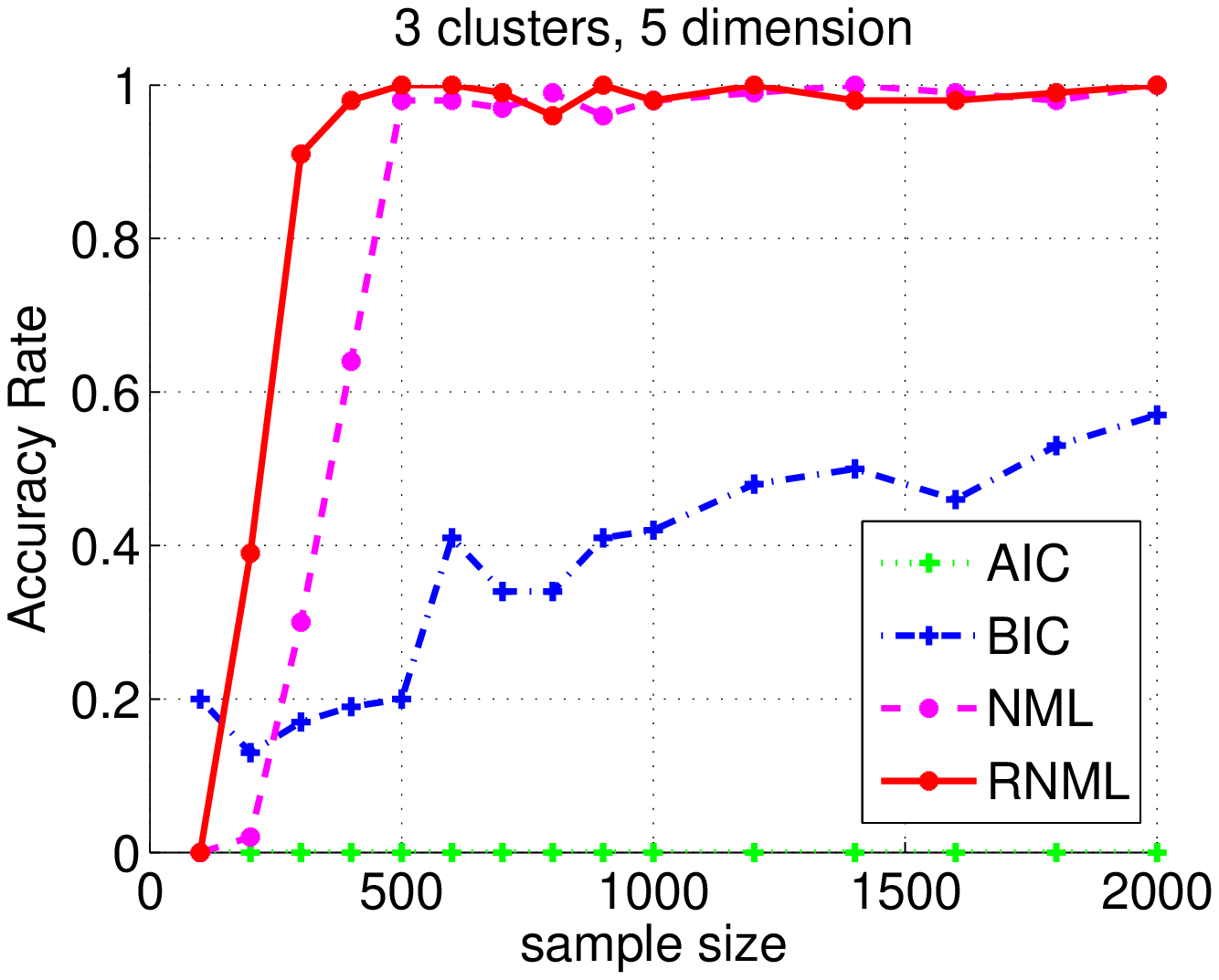} 
\caption{Accuracy Rates} \label{fig:d10K3}
\end{minipage}
\hspace{10pt}
\begin{minipage}{.27\linewidth }
\includegraphics[width=\linewidth,clip ]{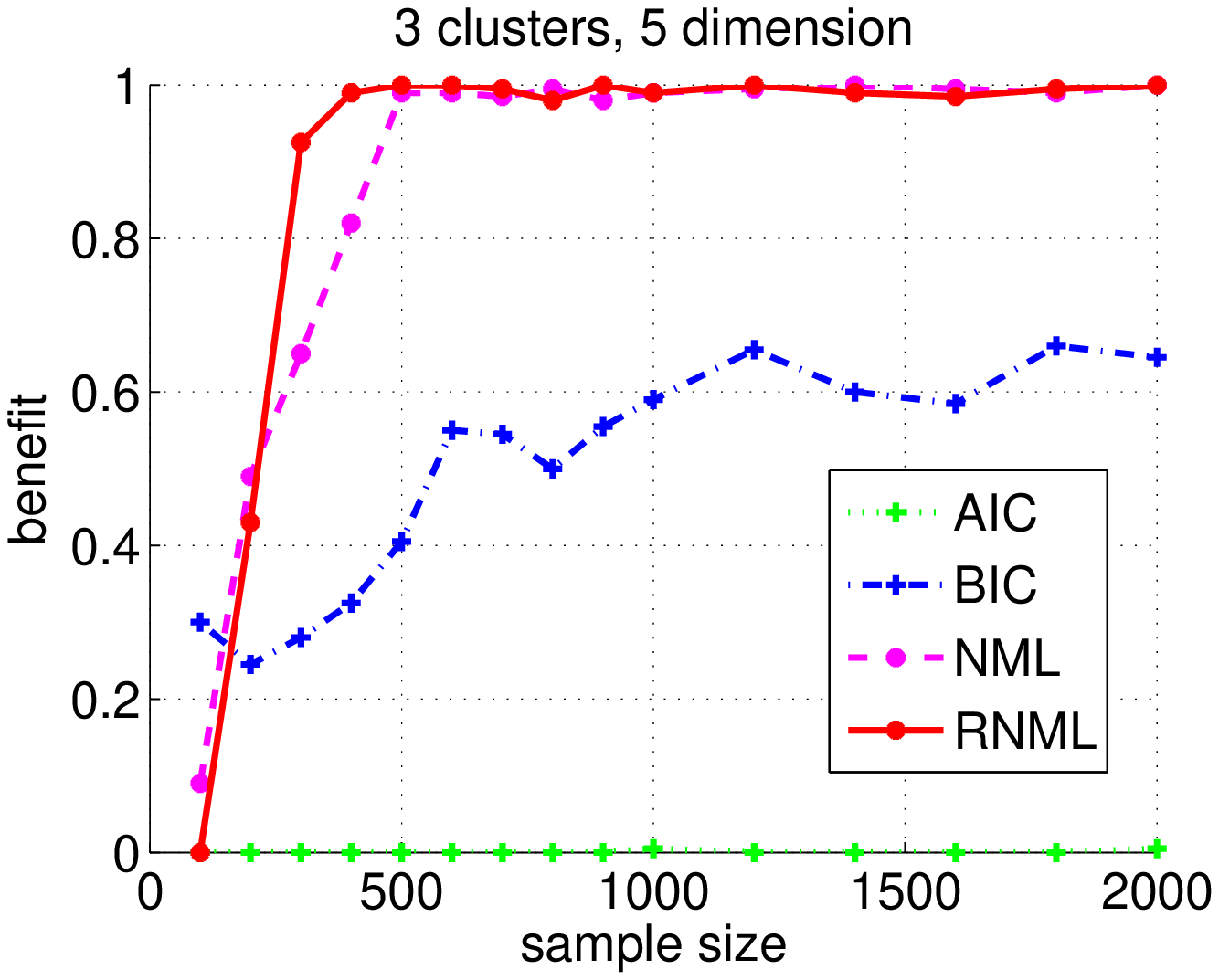} 
\caption{Benefit} \label{fig:d10K3benefit}
\end{minipage}
\hspace{10pt}
\begin{minipage}{.30\linewidth}
\includegraphics[width=\linewidth,clip ]{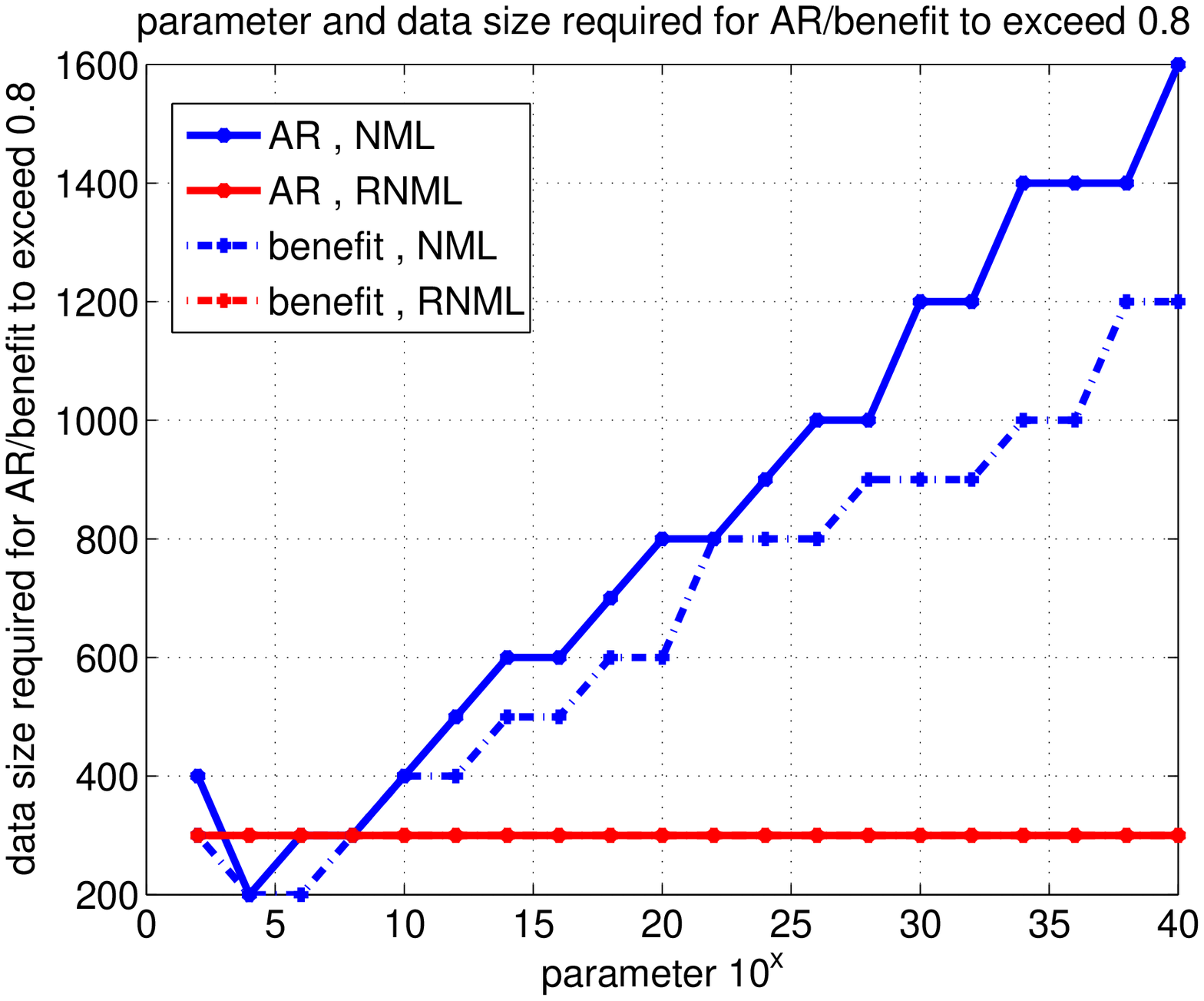} 
\caption{Data Size Required for Accuracy Rate/benefit to Exceed $0.8$ vs Parameter Values} \label{fig:paradata_prob}
\end{minipage}
\end{figure*}

Fig. \ref{fig:d10K3} and Fig. \ref{fig:d10K3benefit} show graphs of 
accuracy rates and benefit vs data size for the case where 
the data dimension was $m=5$ and the true number of clusters was $K=3$.
Here we set $T=2$ in the calculation of $B(K)$ in (\ref{eqn:benefit}).

We see from these results that RNML achieved the highest identification probability, the highest benefit, and the fastest rate of convergence among all of the criteria: AIC, BIC, NML, and RNML.
Specifically this was the case when the data size was not so large.
This implies that RNML was effective as a criterion for selecting an optimal number of clusters even when the data size
was relatively small.

Table \ref{tab:dK_RNML}, \ref{tab:dK_NML}, \ref{tab:dK_AIC}, and \ref{tab:dK_BIC} show the results on benefit obtained by varying the data dimension and the true number of clusters, 
where each numerical value in Tables indicates the least data size required for benefit to exceed $0.8$.
Here Inf shows that benefit did not exceed $0.8$.

We see from these results that for most of pairs of $m$ and $K$,
RNML achieves high benefit with smaller data size than AIC, BIC, and NML.
This implies that the number of clusters estimated by RNML is within $\pm 1$ of the true one  with sufficiently high probability.



\vspace{-3pt}
\begin{table}[!ht]
\begin{minipage}{.48\linewidth}
\begin{center}
\caption{Data Size Required for Benefit to Exceed $0.8$ (RNML)} \label{tab:dK_RNML}
\scalebox{0.7}{
\begin{tabular}{||c||c|c|c|c||}
\hline\hline
m $\backslash$ K & 3 & 4 & 5 & 6 \\
\hline\hline
3 & 300 & 1500 & Inf & Inf \\
\hline
4 & 300 & 300 & Inf & Inf \\
\hline
5 & 300 & 300 & 500 & Inf \\
\hline
6 & 300 & 400 & 600 & 800 \\
\hline\hline
\end{tabular}
}
\end{center}
\end{minipage}
\begin{minipage}{.48\linewidth}
\begin{center}
\caption{Data Size Required for Benefit to Exceed $0.8$ (NML)} \label{tab:dK_NML}
\scalebox{0.7}{
\begin{tabular}{||c||c|c|c|c||}
\hline\hline
m $\backslash$ K & 3 & 4 & 5 & 6 \\
\hline\hline
3 & 600 & 2000 & 5000 & Inf \\
\hline
4 & 600 & 1000 & Inf & Inf \\
\hline
5 & 800 & 1000 & 1500 & Inf \\
\hline
6 & 600 & 1200 & 1500 & 2000 \\
\hline\hline
\end{tabular}
}
\end{center}
\end{minipage}

\vspace{3pt}

\begin{minipage}{.48\linewidth}
\begin{center}
\caption{Data Size Required for Benefit to Exceed $0.8$ (AIC)} \label{tab:dK_AIC}
\scalebox{0.7}{
\begin{tabular}{||c||c|c|c|c||}
\hline\hline
m $\backslash$ K & 3 & 4 & 5 & 6 \\
\hline\hline
3 & Inf & Inf & Inf & Inf \\
\hline
4 & Inf & Inf & Inf & Inf \\
\hline
5 & Inf & Inf & Inf & Inf \\
\hline
6 & Inf & Inf & Inf & Inf \\
\hline\hline
\end{tabular}
}
\end{center}
\end{minipage}
\begin{minipage}{.48\linewidth}
\begin{center}
\caption{Data Size Required for Benefit to Exceed $0.8$ (BIC)} \label{tab:dK_BIC}
\scalebox{0.7}{
\begin{tabular}{||c||c|c|c|c||}
\hline\hline
m $\backslash$ K & 3 & 4 & 5 & 6 \\
\hline\hline
3 & 800 & 1000 & Inf & Inf \\
\hline
4 & 2000 & Inf & Inf & Inf \\
\hline
5 & Inf & Inf & Inf & Inf \\
\hline
6 & Inf & Inf & Inf & Inf \\
\hline\hline
\end{tabular}
}
\end{center}
\end{minipage}
\end{table}
\normalsize

\vspace{-10pt}

\subsection{Dependency of NML and RNML on Parameters} \label{subsec:paraDepend}

Fig.\ref{fig:paradata_prob} 
shows graphs of least data size required for accuracy rate and benefit to achieve $80\%$ and $0.8$ versus parameter values, respectively.
We define parameter $\theta$ as $\theta = R_2/R_1 = \lambda_2/\lambda_1$ in RNML, and $\theta = R = {\lambda_{\rm min}^{(j)}}^{-m}$ in NML.
We see that the RNML do not depend on parameter values more than NML.
It implies that the dependency of RNML on parameter values  is much less than that of NML.


\section{Conclusion}

We have proposed a general method for computing the NML code-length for the exponential family.We have developed it by generalizing the existing method for restricting the data domain so that 
the NML code-length does not diverge.
We have specifically focused on Gaussian distributions and GMMs to 
propose a new efficient method for 
computing the RNML for them. We have developed it by extending Rissanen's renormalizing technique into multi-variate Gaussian distributions. 
We have applied this method to the clustering issue, in which we have selected the optimal number of clusters on the basis of the RNML code-length.
We have empirically demonstrated using artificial data that 
our method makes the estimate of the number of clusters converge significantly faster to the true one than AIC, BIC, and NML.



\begin{thebibliography}{10}

\bibitem{AIC}
H.~Akaike.
\newblock A new look at the statistical model identification.
\newblock {\em IEEE Trans. on Automatic Control}, 19(6):716--723, Dec. 1974.

\bibitem{dlr}
A.~P. Dempster, N.~M. Laird, and D.~B. Rubin.
\newblock Maximum likelihood from incomplete data via the em.
\newblock {\em J.Royal Staitst. Soc.B}, 39:1--38, 1977.

\bibitem{Giu}
C.~D. Giurc$\breve{a}$neanu, S.~A. Razavi, and A.~Liski.
\newblock Variable selection in linear regression: Several approaches based on
  normalized maximum likelihood.
\newblock {\em Signal Processing}, 91(8), March 2011.

\bibitem{Grunwald2007}
P.~D. Gr$\ddot{\rm u}$nwald.
\newblock {\em The Minimum Description Length Principle}.
\newblock MIT Press, Cambridge, June 2007.

\bibitem{HiraiNML}
S.~Hirai and K.~Yamanishi.
\newblock Efficient computation of normalized maximum likelihood coding for
  gaussian mixtures with its applications to optimal clustering.
\newblock {\em The IEEE ISIT}, pages 1031--1035, 2011.

\bibitem{petri1}
P.~Kontkanen and P.~Myllym\"{a}ki.
\newblock A linear time algorithm for computing the multinomial stochastic
  complexity.
\newblock {\em Information Processing Letters}, 103:227--233, 2007.

\bibitem{petri2}
P.~Kontkanen and P.~Myllym\"{a}ki.
\newblock An empirical comparison of nml.
\newblock {\em Proceedings of the 2008 International}, pages 125--131, 2008.

\bibitem{bibli_10}
J.~Rissanen.
\newblock Fisher information and stochastic complexity.
\newblock {\em IEEE Trans. on Information Theory}, 42(1):40--47, January 1996.

\bibitem{Denoise}
J.~Rissanen.
\newblock {MDL} denoising.
\newblock {\em IEEE Trans. on Information Theory}, 46(7):2537--2543, November
  2000.

\bibitem{infcomp}
J.~Rissanen.
\newblock {\em Information and Complexity in Statistical Modeling}.
\newblock Springer, 2007.

\bibitem{BIC}
G.\ Schwarz.
\newblock Estimating the dimension of a model.
\newblock {\em Annals of Statistics 6 (2)}, pages 461--464, 1978.

\bibitem{shtarkov}
Shtarkov Yu.~M.
\newblock Universal sequential coding of single messages.
\newblock {\em Problems of Information Transmission}, 23(3):3--17,
  July-September 1987.

\end{thebibliography}


\end{document}